# A Categorical Approach for Recognizing Emotional Effects of Music

Mohsen Sahraei Ardakani [1] and Ehsan Arbabi [2]

School of Electrical and Computer Engineering, College of Engineering, University of Tehran, Iran

[1] m.ardakani@alumni.ut.ac.ir, [2] earbabi@ut.ac.ir

*Abstract* – Recently, digital music libraries have been developed and can be plainly accessed. Latest research showed that current organization and retrieval of music tracks based on album information are inefficient. Moreover, they demonstrated that people use emotion tags for music tracks in order to search and retrieve them. In this paper, we discuss separability of a set of emotional labels, proposed in the categorical emotion expression, using Fisher's separation theorem. We determine a set of adjectives to tag music parts: happy, sad, relaxing, exciting, epic and thriller. Temporal, frequency and energy features have been extracted from the music parts. It could be seen that the maximum separability within the extracted features occurs between relaxing and epic music parts. Finally, we have trained a classifier using Support Vector Machines to automatically recognize and generate emotional labels for a music part. Accuracy for recognizing each label has been calculated; where the results show that epic music can be recognized more accurately (77.4%), comparing to the other types of music.

## 1 Introduction

People listen to different kinds of music in their daily activities. It has been proved that music can evoke emotion in listeners and change their mood [1]. Music libraries with the help of internet and high quality compressed formats such as MP3 enable people to access a wide variety of music on a daily basis [2]. The increasing size of these libraries makes their organization based on album information such as album name, artist, and composer inefficient. Organization of these libraries must develop in a way that provides easy access to the data and meta-data [3, 4]. Former studies showed that 28.2% of users utilize emotional labels for archiving and searching music [5, 6]. Human emotion system is subject of many scholarly studies in different areas [7]. Emotions are analyzed in three phases: emotion expressed, emotion perceived and emotion evoked. Emotion perceived is considered to be subject-independent [8] and we focus on this functionality of the music.

Human verbal language has inherent ambiguity [9]. Psychological studies illustrated that people can successfully recognize their emotions but fail to describe them [10]. This ambiguity causes serious problems when it comes to different adjectives with similar meaning. Some research proposed using a set of basic emotions for emotion description. They put adjectives that express an emotion with similar meaning in a same cluster [1, 11]. In this case as the number of basic emotions increases the accuracy of emotion detection decreases. However, a limited number of basic emotions do not provide desired resolution in emotion description [12].

The other issue is the subjectivity of the emotions evoked. Muyuan et al. concluded that cultural background, age, gender, personality, etc. affects human-music emotional interaction [13]. Current solution to this issue is to hold on to those music tracks that result in similar emotion responses in people with different situations [5]. Considering this solution, we limit the under-study music to those for which the emotional content can be obtained apart from subjectivity issues.

Recently, much research has been published in music emotion recognition, where some of them are only applied to a specific music genre. The outline of these studies consists of steps: 1- data collection 2-data processing and feature extraction 3- machine learning algorithm. In these works, data collection is done individually because depending on utilized emotion taxonomy, appropriate data collection scenario is adopted and a common data set cannot be used as reference [14]. Nevertheless, there are some rules to follow. All music parts are altered to the standard form. Some measures are adopted in a way that subjects' memory or album effect does not affect their assessment [15].

Jun et al. modified Thayer's Arousal-Valence model and categorized emotions in eleven classes [16]. They concluded that arousal level of a music part highly correlates with the intensity feature set, and rhythm feature set correlates with valence level. In a similar work, Lu et al. divided Arousal-Valence plane into four categories and extracted low level features in order to find relationship between feature sets and arousal and valence levels of music parts [17]. Yang and Chen expressed emotions as points in Arousal-Valence plane [18]. Although they did not encounter the ambiguity issues in describing emotions with verbal language, the problem remains unsolved because it fails to provide verbal description.

Our work, which was basically done in 2013 in the School of Electrical and Computer Engineering at University of Tehran, strives for presenting a computational model of music emotion by extracting different sets of features, including timbre, harmony, rhythm and energy. These feature sets tend to represent emotional content of the music [11, 13, 19]. Objective here is to investigate relation between emotional content of the music and the extracted feature sets. In this paper, we exploit a set of adjectives covering Thayer's Arousal-Valence plane and some other adjectives to cover third dimension of extended version of this emotion taxonomy. Including adjectives related to stance or dominance helps subjects to describe their emotion with a better resolution. With the use of Fisher's Separation Theorem, we discuss efficiency of the adjective set and after that by using Support Vector Machines (SVMs), we train a classifier for automatic recognition of emotional labels.

The rest of this paper is structured as follows. In section 2 an overview of emotion description is introduced. In section 3 the extracted feature sets are presented. In section 4 the performed experiment is reported. In section 5 the efficiency of the proposed six labels is investigated and finally section 6 concludes this paper.

## 2 Music Emotion Taxonomy

Psychologists usually use verbal assessment of subjects in emotion recognition studies [7]. In the categorical emotion recognition, adjectives expressing emotions are categorized in a specific number of clusters [20]. Although categorical approach to emotion expression provides verbal description of emotions; it fails to differentiate synonymous adjectives as they all go in the same cluster but offer different meanings literally. Three basic introduced factors in dimensional approach make it possible to locate all emotions in space. According to K. R. Scherer these basic factors are Arousal, Valence and Dominance [21]. However, Thayer's Arousal-Valence is the most common metric in music emotion recognition [7]. In Thayer's model Valence varies from negative to positive and Arousal from calm to excited [22]. In this paper considering benefits of a limited number of factors and demanding verbal descriptive labels to provide meta-data, we propose using a set of adjectives covering three-dimensional space of emotions, we furthermore discuss its efficiency using Fisher's Separation theorem.

## 3 Feature Extraction

The objective here is to extract features that can present a computational model of acoustic cues. Specific patterns of these features modulate different emotions. Although the relation between the emotion evoked and some of these features are predictable; but in this work, we do not stick out to low level features in order to achieve a better accuracy. Different sets of features are extracted representing different characteristics of the music cue. Intensity features represent energy content of the music cue. Timbre features are considered to represent spectral shape of the music cue. Mel Frequency Cepstral Coefficients (MFCCs) represent effect of frequency content of music on human hearing system. The other sets frame regularity, mode and temporal shape of the music signal.

### *3.1 Intensity Features*

Intensity features represent the energy content of music signals. Intensity features are calculated uniquely using frequency domain. Their relation with different arousal levels is predictable [17]. Intensity features are calculated using Fast Fourier Transform (FFT) of acoustic signal in consecutive frames of the music part. Using FFT coefficients, intensity in frequency sub-bands is calculated. Sub-bands are determined in equation 1, in which $f_0$ is the sampling frequency. Equation 2 defines intensity of $n^{th}$ frame where $A(n, k)$ is absolute value of $k^{th}$ FFT coefficient of $n^{th}$ frame. Equation 3 is the ratio of Intensity in $i^{th}$ sub-band (between $L_i$ and $H_i$) of $n^{th}$ frame to its total intensity. Average and standard deviation of energy sequence of each frame represent the regularity of the acoustic signal [16]. These metrics are shown in equation 4 and 5 ($x[n]$ is an input discrete signal).

$$\left[0, \frac{f_0}{2^n}\right), \left[\frac{f_0}{2^n}, \frac{f_0}{2^{n-1}}\right), \left[\frac{f_0}{2^{n-1}}, \frac{f_0}{2^{n-2}}\right), \dots, \left[\frac{f_0}{2^2}, \frac{f_0}{2^1}\right) \quad (1)$$

$$I(n) = \sum_{k=0}^{\frac{f_0}{2}} A(n, k) \quad (2)$$

$$D_i(n) = \frac{1}{I(n)} \sum_{k=L_i}^{H_i} A(n, k) \quad (3)$$

$$AE\{x[n]\} = \frac{1}{N} \sum_{i=0}^{N} |x[n]^2| \quad (4)$$

$$\sigma\{AE\{x[n]\}\} = \sqrt{\frac{1}{N}\sum_{i=0}^{N}[x^2[n] - AE\{x[n]\}]^2} \qquad (5)$$

### *3.2 Timbre Features*

This group of features represents spectral properties of the acoustic signal and can be extracted using different methods. Equation 6 defines the centroid frequency of $n^{th}$ frame. Roll-off frequency is calculated in equation 7 where R[n] is roll-off frequency of $n^{th}$ frame. Spectral flux is defined by equation 8, which represents the intensity of spectral density variations in adjacent frames [23]. Average and standard deviation of these parameters can be used as timbre features.

$$C[n] = \frac{\sum_{k=0}^{\frac{f_0}{2}} A(n,k)*k}{\sum_{k=0}^{\frac{f_0}{2}} A(n,k)} \qquad (6)$$

$$\sum_{k=0}^{R[n]} A(n,k) = 0.85 \times \sum_{k=0}^{\frac{f_0}{2}} A(n,k) \qquad (7)$$

$$F[n] = \sum_{k=0}^{\frac{f_0}{2}} [A(n,k) - A(n-1,k)]^2 \qquad (8)$$

### *3.3 Mel-Frequency Cepstral Coefficients (MFCCs)*

MFCCs are calculated considering human hearing system, which represent the frequency content of acoustic cues [23]. In this paper average and standard deviation of the first 20 coefficients in consecutive frames are utilized.

### *3.4 Rhythm Features*

Rhythm is one of the most basic features of the music cues. Different rhythms make listener experience various emotional states [16]. Moreover, beat and tempo are extracted from rhythm histogram. They are highly correlated with the arousal level of music signal. Rhythm is defined as music pattern in time. We have estimated the rhythmicity of music signal by using MIRtoolbox of MATLAB [24].

### *3.5 Harmony Features*

Features related to *mode* are used to achieve different emotional constructions in music science. Here, *mode* is defined as the difference between the strongest minor key and the strongest major key, which can be a robust factor in valence determination. Inharmonicity is the number of partials that are not multiple of the fundamental frequency. We have used MIRtoolbox for estimating inharmonicity of audio signal and the numerical value of modality [24].

### *3.6 Temporal Features*

One of the temporal features of acoustic cues is the zero-crossing rate. Zero-crossing is calculated using Equation 9 [23]. Average and standard deviation of zero-crossing in frames are used as features.

$$ZC[n] = \frac{1}{2}\sum_{i=1}^{N}|sgn(x[n,i]) - sgn(x[n, i-1])| \quad (9)$$

Autocorrelation of music signal can be used as a measure of uniformity. In this research first 13 coefficients of autocorrelation of music signal are used for this purpose.

## 4 Experiment

A large set of instrumental music tracks (without vocals) were collected covering different music genres. In music selection, due to subjectivity issues of the emotion evoked, it was tried to select music tracks having similar emotion response for different people. To avoid the album effect and complexities associated with the lyrics, music tracks with singing were excluded. 15-second parts from 93 remaining music tracks were cut manually with the purpose of avoiding music emotion variation. In order to include all emotion classes, emotional labels of *last.fm* website were used [25]. In preprocessing, music parts were altered to a standard form of 16-bit precision, mono-channel wav format and re-sampled to 22'025 Hz. Maximum sound volumes were fixed to a constant value for all the music parts. For extracting the features, number of frequency sub bands and number of time frames have been set to 10 and 124, respectively.

18 subjects assessed their evoked emotion after listening to the music parts. Their evoked emotion was evaluated using the six emotional labels. In order to achieve the desired accuracy, evaluations for music parts calling up a memory were discarded. Labels supported by the majority of the subjects were assigned to music parts and considered to express the emotional content of music parts.

## 5 Results and Discussion

Using Fisher's separation theorem, pairwise separability of labels was calculated for all the features. From the six labels, two of them are mostly related to the valence level (happy, sad), two of them are mostly related to the arousal level (relaxing, exciting) and the other two mostly describe dominance factor emotion (epic, thriller). The highest separability of two labels indicates which feature is the most decisive. It is necessary to mention that low separability of pairs can be interpreted as paucity of music set or correlation between labels. One of the innovations in this work was adding labels to cover third dimension of Scherer model. Note that other studies used two-dimensional Arousal-Valence plane and one of the issues mentioned before was failing in emotion description. The proposed adjective set here provide desired resolution and help subjects to describe their emotion evoked more accurately. As it is demonstrated in Table 1 epic label, which has the highest separability comparing to the other labels, indicates its efficiency to provide desired meta-data. Fisher's separability for feature *f* is shown in Equation 10. In this equation, $\mu_i$ and $\sigma_i$ are mean and standard deviation of a feature (*f*) extracted from the data with label i, respectively.

$$\text{Separability}(f) = \frac{(\mu_1 - \mu_2)^2}{\sigma_1^2 + \sigma_2^2} \quad (10)$$

Table 1. Maximum separability

| Label | Happy | Sad | Relaxing | Exciting | Epic | Thriller |
|---|---|---|---|---|---|---|
| **Happy** | - | 1.46 | 0.89 | 0.33 | 1.29 | 1.25 |
| **Sad** | 1.46 | - | 0.06 | 0.83 | 1.62 | 0.35 |
| **Relaxing** | 0.89 | 0.06 | - | 0.78 | 2.08 | 0.47 |
| **Exciting** | 0.33 | 0.83 | 0.78 | - | 0.44 | 0.54 |
| **Epic** | 1.29 | 1.62 | 2.08 | 0.44 | - | 1.83 |
| **Thriller** | 1.25 | 0.35 | 0.47 | 0.54 | 1.83 | - |

The other result to be noted is the most determinant feature set in each dimension. The most determinant feature in valence dimension is rhythm. As referenced before valence level cannot be determined without the use of high-level features such as rhythm. On the other side, the most decisive feature in arousal dimension is related to intensity and MFCC. Maximum separability and the feature group corresponding to this maximum separability are reported in Table 1 and Table 2, respectively.

Table 2. Features causing the maximum separability

| Label | Happy | Sad | Relaxing | Exciting | Epic | Thriller |
|---|---|---|---|---|---|---|
| **Happy** | - | Rhythm | Rhythm | MFCC | MFCC | Rhythm |
| **Sad** | Rhythm | - | Rhythm | MFCC | Rhythm | MFCC |
| **Relaxing** | Rhythm | Rhythm | - | MFCC | MFCC | MFCC |
| **Exciting** | MFCC | MFCC | MFCC | - | Intensity | Rhythm |
| **Epic** | MFCC | Rhythm | MFCC | Intensity | - | Rhythm |
| **Thriller** | Rhythm | MFCC | MFCC | Rhythm | Rhythm | - |

In Table 3 average and standard deviation of maximum separability values for each label is reported. The results depict that the epic label besides providing the description of the third dimension of emotions, has the highest average among the labels. In addition to providing verbal description and better resolution in emotion description, from Table 3 it is construed that the epic label is highly separable in the space of features.

Table 3. Average and standard deviation of maximum separability for each label

| Label | Happy | Sad | Relaxing | Exciting | Epic | Thriller |
|---|---|---|---|---|---|---|
| **Average** | 1.04 | 0.86 | 0.86 | 0.58 | 1.45 | 0.89 |
| **STD** | 0.45 | 0.67 | 0.76 | 0.21 | 0.64 | 0.63 |

A classifier was trained using Support Vector Machines in order to recognize music label automatically. In each turn one music part was considered as the test data and all the remaining music parts were included in the train data. In the next turn, another music part was considered as the test data. Continuing this process for all the music parts, accuracy of automatic music label recognition was calculated (see Table 4). The maximum accuracy happens when recognizing Epic and Happy music (77.4% and 76.3%). On the other hand, the minimum accuracy is related to recognizing Relaxing music (40.9%). It should be noticed that in a random recognition system, the accuracy is about 16.7%, which is much lower than 40.9%.

Table 4. Accuracy of the recognized labels

| Label | Happy | Sad | Relaxing | Exciting | Epic | Thriller |
|---|---|---|---|---|---|---|
| **Accuracy (%)** | 76.3 | 53.8 | 40.9 | 64.5 | 77.4 | 67.7 |

## 6 Conclusion

In the digital age, organization and retrieval of data should be in a way that provides proper access to large-scale digital libraries. Emotional tags facilitate obtaining demanded meta-data. In order to automatically generate emotional labels, it is fundamental to possess an emotional label set expressing emotional states and avoiding misapprehension and complexity. In our work, a set of labels were proposed and its efficiency was investigated. Using third dimension of emotion space enables users to succeed in describing their emotion. The important achievement is that the proposed adjectives set in addition to providing verbal description and covering three-dimensional emotion space, shows the desired efficiency. The proposed emotion taxonomy in this article, included epic label to enable users evaluate stance feature of emotion content of music parts. The epic label in addition to providing verbal description of the stance quality of emotions is highly distinguished in feature space. By using a classifier, proper accuracy was achieved in automatic recognition of emotional labels. In future studies, by utilizing proper music set and using high-level features, higher accuracy in determination of emotional content of music may be obtained.

## Acknowledgement

The authors would like to thank Mostafa Sahraei Ardakani for his assistant during editing the manuscript.

## References

[1] Feng, Y., Zhuang, Y., Pan, Y.: Popular music retrieval by detecting mood. Proceedings of the 26th annual international ACM SIGIR conference on Research and Development in Information Retrieval, pp.375-376, 2003.

[2] Wieczorkowska, A., Synak, P., Ras, Z.: Multi-label classification of emotions in music. In Klopotek, M., Wierzchon, S., Trojanowski, K., eds.: Intelligent Information Processing and Web Mining. Springer Berlin Heidelberg, 307-315, 2006.

[3] Lee, C., Narayanan, S.: Toward detecting emotions in spoken dialogs. IEEE Transactions on Speech and Audio Processing 13(2), 293-303, 2005.

[4] Huron, D.: Perceptual and cognitive applications in music information retrieval. Perception 10(1), 83-92, 2000.

[5] Agresti, A.: Categorical data analysis. John Wiley, New Jersey, 2002.

[6] Laurier, C., Sordo, M., Serra, J., Herrera, P.: Music mood representations from social tags. International Society for Music Information Retrieval (ISMIR) Conference, pp.381-386, 2009.

[7] Juslin, P., Sloboda, J.: Music and Emotion: Theory and Research. Oxford University Press, New York, USA, 2001.

[8] Gabrielsson, A.: Emotion perceived and emotion felt: Same or different? Musicae Scientiae 5(1), 123–147, 2002.

[9] Hersh, H., Caramazza, A.: A fuzzy set approach to modifiers and vagueness in natural language. Journal of Experimental Psychology: General 105(3), 251-276, 1976.

[10] Posner, J., Russell, J., Peterson, B.: The circumplex model of affect: An integrative approach to affective neuroscience. Development and Psychopathology 17(3), 715–734, 2005.

[11] Li, T., Ogihara, M.: Detecting emotion in music. International Society for Music Information Retrieval (ISMIR) Conference, pp.239-240, 2003.

[12] van de Laar, B.: Emotion detection in music, a survey. Twente Student Conference on IT, 700, 2006.

[13] Muyuan, W., Naiyao, Z., Hancheng, Z.: User-adaptive music emotion recognition. 7th IEEE International Conference on Signal Processing (ICSP'04), p.1352–1355, 2004.

[14] Lee, D., Yang, W.-S.: Disambiguating music emotion using software agents. International Society for Music Information Retrieval (ISMIR) Conference, pp.218-223, 2004.

[15] Kim, Y., Williamson, D., Pilli, S.: Towards quantifying the "album effect" in artist identification. International Society for Music Information Retrieval (ISMIR) Conference, pp.393-394, 2006.

[16] Jun, S., Rho, S., Han, B.-j., Hwang, E.: A fuzzy inference-based music emotion recognition system. 5th International Conference on Visual Information Engineering, pp.673-677, 2008.

[17] Lu, L., Liu, D., Zhang, H.-J.: Automatic Mood Detection and Tracking of Music Audio Signals. IEEE Transactions on Audio, Speech, and Language Processing 14(1), 5-18, 2006.

[18] Yang, Y., Chen, H.: Searching music in the emotion plane. IEEE MMTC E-Letter, 2009.

[19] Katayose, H., Imai, M., Inokuchi, S.: Sentiment extraction in music. 9th IEEE International Conference on Pattern Recognition, p.1083–1087, 1998.

[20] Juslin, P., Laukka, P.: Expression, perception, and induction of musical emotions: A review and a questionnaire study of everyday listening. Journal of New Music Research 33(3), 217-238, 2004.

[21] Scherer, K.: Which emotions can be induced by music? what are the underlying mechanism? and how can we measure them. Journal of New Music Research 33(3), 239–251, 2004.

[22] Kim, J., Andre, E.: Emotion recognition based on physiological changes in music listening. IEEE Transactions on Pattern Analysis and Machine Intelligence 30(12), 2067-2083, 2008.

[23] Tzanetakis, G., Cook, P.: Musical genre classification of audio signals. IEEE Transactions on speech and audio processing 10(5), 293-302, 2002.

[24] Lartillot, O., Toiviainen, P.: A Matlab Toolbox for Musical Feature Extraction from Audio. International Conference on Digital Audio Effects, Bordeaux, 2007.

[25] In: last.fm. Available at: http://www.last.fm.